%% file: main.tex
\documentclass[10pt,twocolumn,letterpaper]{article}
\usepackage[pagenumbers]{cvpr}

\input{preamble}
\definecolor{cvprblue}{rgb}{0.21,0.49,0.74}
\usepackage[breaklinks,colorlinks,allcolors=cvprblue]{hyperref}
\input{cmd/math}

\input{cmd/packages}

\title{One-to-Many: Diverse and Bidirectional\\Image-to-Image Translation via Rectified Flow}
\input{authors}
\date{}
\hypersetup{
  pdftitle={One-to-Many: Diverse and Bidirectional Image-to-Image Translation via Rectified Flow},
  pdfsubject={Image-to-image translation via rectified flow}
}

\begin{document}
\raggedbottom
\maketitle
\input{sec/0_abstract}
\input{sec/1_intro}
\input{sec/2_relatedwork}
\input{sec/3_preliminary}
\input{sec/4_proposedmethod}
\flushbottom
\input{sec/5_experiments}
\input{sec/6_conclusion}
\par\pagebreak
\raggedbottom
\input{sec/8_llm_usage}
{
    \small
    \bibliographystyle{ieeenat_fullname}
    \bibliography{main}
}

% Supplementary material is included in the same preprint.
\input{sec/x_supp}
\end{document}

%% file: preamble.tex
\usepackage[expansion=false]{microtype}

%% file: cmd/math.tex
\newcommand{\cmark}{\ding{51}} % check mark

%% file: cmd/packages.tex
\usepackage{bm}
\usepackage{multirow}
\usepackage[table]{xcolor}
\usepackage{pifont}

%% file: authors.tex
\author{Sho Shimotsumagari\thanks{Current affiliation: CyberAgent, Inc.}
\quad Seiichi Uchida \quad Shumpei Takezaki\\
Kyushu University, Fukuoka, Japan\\
{\tt\small\{sho.shimotsumagari, uchida, shumpei.takezaki\}@human.ait.kyushu-u.ac.jp}}

%% file: sec/0_abstract.tex
\begin{abstract}
Many paired image-to-image translation tasks admit multiple valid target appearances, while training provides only one target per source. We propose One-to-Many Flow (O2MFlow), a latent-conditioned Rectified Flow (RF) model for one-to-many generation and backward translation (target$\rightarrow$source) within a single transport framework. Conditioning the RF dynamics on an appearance latent turns transport into a family of latent-indexed bijections: a fixed latent gives a mapping that is inverted by time-reversed integration, and sampled latents produce diverse outputs for the same input. We formalize reusable appearance variations as Shared Latent Transport and show that Kullback--Leibler (KL) regularization directly controls the source--latent distribution gap between training and generation. Across three benchmarks, O2MFlow offers dataset-dependent forward diversity--fidelity trade-offs and the lowest backward LPIPS-Dist among evaluated methods, with the lowest backward FID on two benchmarks.
\end{abstract}

%% file: sec/1_intro.tex
\section{Introduction}
\label{sec:intro}

In appearance-ambiguous tasks, paired image-to-image translation is one-to-many, yet each training source has only one observed target. The task maps a source-domain image to a target domain while preserving task-relevant structure, as in edge-to-photo synthesis and semantic label-to-image generation~\cite{isola2017pix2pix}. The input, however, constrains geometry but leaves appearance underdetermined~\cite{zhu2017bicyclegan}: the same edge map corresponds to many plausible photos with different colors, materials, or lighting. A useful translator must therefore produce diverse outputs that keep the input structure, and it must learn these variations without ever observing two valid targets for the same input.

\input{fig/img2img}
\input{fig/rf_vs_ours}

Many applications also require \emph{backward translation} (target$\rightarrow$source), e.g., extracting a structural representation from a real target-domain image for editing, verification, or analysis. A separately trained backward predictor is not the inverse of the forward mapping, so the two directions can disagree on the same image pair. Backward translation should therefore invert the forward model.

Existing translation frameworks lack a unified mechanism for controllable diverse generation and invertible transport, as summarized in Fig.~\ref{fig:img2img}. Fig.~\ref{fig:img2img}(a) shows conventional single-output models such as pix2pix~\cite{isola2017pix2pix}, which learn non-invertible mappings and yield a single output per input. Fig.~\ref{fig:img2img}(b) illustrates invertible transport models based on Rectified Flow (RF)~\cite{liu2023rectifiedflow}. Because inversion is obtained by time-reversed integration of the same ordinary differential equation (ODE), these models naturally support backward translation; however, standard RF models a one-to-one mapping and cannot represent diverse outputs for the same input. In contrast, Fig.~\ref{fig:img2img}(c) represents latent-variable approaches such as BicycleGAN~\cite{zhu2017bicyclegan}, which generate diverse outputs but map latents to outputs in a single feedforward pass and provide no inverse from target to source. Diffusion bridges~\cite{li2023bbdm,kieu2025bdbm} draw diverse samples, even in both directions, but do not reverse the forward map using a shared appearance latent. Fig.~\ref{fig:img2img}(d) depicts the desired setting: a single framework that supports both one-to-many forward translation and backward translation.

In this paper, we propose One-to-Many Flow (O2MFlow), a latent-conditioned RF framework that achieves one-to-many generation and backward translation with a single transport model trained from one observed target per source. Conditioning RF dynamics on an appearance latent turns a deterministic ODE transport into a family of latent-indexed bijections. Sampling different latents from a prior yields diverse outputs for the same input, while a fixed latent defines an invertible transport (Sec.~\ref{sec:standardRF}), so backward translation is time-reversed integration of the same dynamics rather than a separate generator.

% Balance page 2 through local line breaking without changing the text.
{\looseness=2
O2MFlow learns from single pairs by conditioning the field on latents inferred from the observed target. Target-encoded latents can convey appearance information, and the shared field allows them to be reused across inputs. We formalize valid, diverse outputs with cross-source consistency as Shared Latent Transport and examine this behavior empirically. Training uses target-encoded latents, whereas generation samples the prior. We show that Kullback--Leibler (KL) regularization bounds the resulting source--latent distribution gap, equaling it under the empirical distribution with a unique target per source; this explains its role in aligning training \mbox{and generation.}
\par}

{\looseness=1
Experiments on three paired benchmarks show forward diversity--fidelity trade-offs and the lowest backward LPIPS-Dist among evaluated methods; backward FID is lowest on edges2shoes and edges2handbags. Same-latent round trips assess numerical reversibility separately from real-target recovery.
\par}

Our contributions are summarized as follows:
\begin{itemize}
  \item We propose O2MFlow, a paired image-to-image translation framework that learns one-to-many generation from one observed target per source and performs backward translation by reversing the same fixed-latent ODE transport.
  \item We formalize reusable appearance variation as Shared Latent Transport and show that KL regularization directly controls the source--latent distribution gap between paired training and prior-based generation, equaling it under the empirical distribution with a unique target per source.
  \item We evaluate forward diversity, fidelity, and real-target backward accuracy on three paired benchmarks, with synchronized-latent analysis and component ablations.
\end{itemize}

%% file: fig/img2img.tex
\begin{figure}[t]
  \centering
  \includegraphics[width=\linewidth]{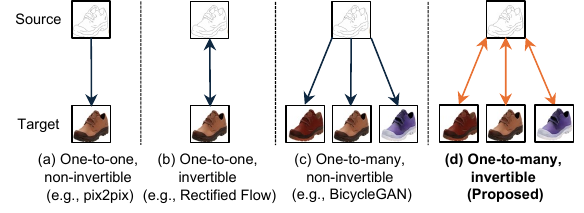}
  \caption{Four representative settings of image-to-image translation. The proposed method corresponds to (d), supporting one-to-many paired translation and backward translation within a single framework.}
  \label{fig:img2img}
\end{figure}

%% file: fig/rf_vs_ours.tex
\begin{figure*}[t]
  \centering
  \includegraphics[width=0.9\linewidth]{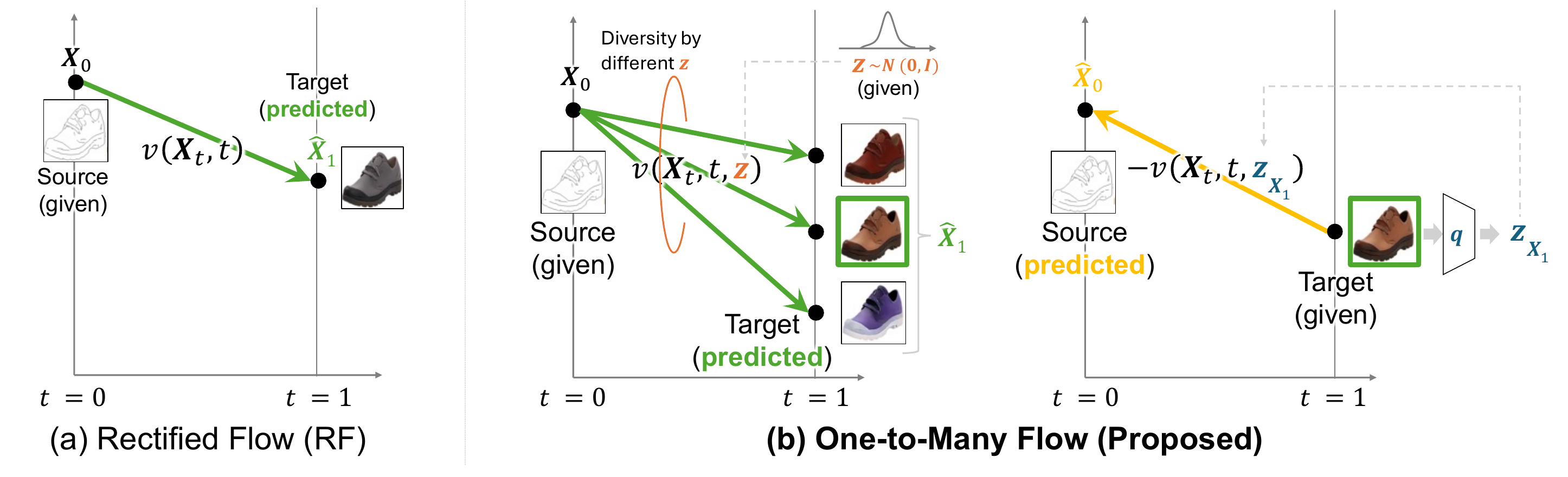}
\caption{(a)~Standard RF. (b)~O2MFlow samples appearance latents for diverse forward translation and infers a latent from the target for backward translation. Here $q$ denotes the target-only encoder and $\mathbf{z}_{\mathbf{X}_1}$ a latent inferred from target $\mathbf{X}_1$; parameter subscripts are omitted. The vertical axis represents image space.}
  \label{fig:RF-vs-ours}
\end{figure*}

%% file: sec/2_relatedwork.tex
\section{Related Work}
\label{sec:related-work}

\paragraph{Multimodal image-to-image translation.}
% Balance page 2 through local line breaking without changing the text.
{\looseness=1
Latent-variable designs make paired image-to-image translation one-to-many, in contrast to single-output conditional mappings such as pix2pix~\cite{isola2017pix2pix}.
BicycleGAN combines a cVAE-GAN branch, which already uses target-image encoding and KL regularization, with a latent-regression branch to encourage diverse outputs~\cite{zhu2017bicyclegan}. Content--style disentanglement frameworks such as MUNIT/DRIT/DRIT++ generate diverse samples by recombining a shared content representation with a stochastic style latent~\cite{huang2018munit,lee2018drit,lee2020dritpp}, and StarGAN v2 scales style-based modulation to multiple target domains~\cite{choi2020starganv2}.
These generators map a latent to an output in a single feedforward pass, without an explicit inverse transport back to the source domain. O2MFlow instead obtains backward translation by reversing the same fixed-latent image transport.
\par}

\paragraph{Diffusion and bridge-based translation.}
{\looseness=1
Diffusion and bridge-based conditional generators achieve strong sample fidelity for image-to-image translation by iteratively refining a noisy variable toward the target domain. Palette provides a unified conditional diffusion framework for a broad range of image-to-image translation tasks~\cite{saharia2022palette}, and bridge formulations such as I$^2$SB, BBDM, and DDBM explicitly construct stochastic processes connecting the source and target distributions~\cite{liu2023i2sb,li2023bbdm,zhou2024ddbm}. Bidirectional bridges such as BiBBDM and BDBM further train a single network for both directions and select the direction at sampling time~\cite{xue2025bibbdm,kieu2025bdbm}. These models draw diversity from sampling noise rather than an explicit appearance latent, and bidirectional sampling does not reverse a fixed-latent transport. O2MFlow instead holds a latent fixed so that backward translation reverses the \mbox{forward transport.}
\par}

\paragraph{Flow Matching and Rectified Flow.}
Continuous-time flow models provide the invertible transport on which O2MFlow builds: under existence and uniqueness conditions, integrating the same dynamics backward in time inverts the forward map, as in neural ODEs and continuous normalizing flows (CNFs)~\cite{chen2018neuralode,grathwohl2019ffjord}.
Flow Matching and RF train such vector fields with regression-style objectives~\cite{lipman2023flowmatching,liu2023rectifiedflow}, and recent work scales RF to higher-resolution synthesis~\cite{esser2024rft} and adapts it to paired image-to-image translation with higher fidelity at lower cost~\cite{iizuka2026i2irfr,mahara2026pairflow}.
Because inversion is implemented as a numerical ODE solve, several studies propose solver/inference-time improvements, stochastic RF variants, and bidirectional training for inversion and editing~\cite{wang2025tamingrf,deng2025fireflow,rout2025rfsde_inversion,dai2026bifm}.
SymmFlow learns symmetric flow-matching dynamics between semantic maps and images for bidirectional synthesis~\cite{caetano2026symmflow} and serves as our closest bidirectional baseline. The paired RF translators start from Gaussian noise with the source as a condition, so they neither transport the source image itself nor invert to it, and SymmFlow has no appearance latent indexing the transport; O2MFlow instead transports the source image along an invertible path indexed by a reusable appearance latent.

Latent-conditioned and variational flow formulations add latent variables to flow models to capture multimodal velocities, straighten trajectories, or improve efficiency~\cite{guo2025vrfm,ma2026straightflows,samaddar2026latentcfm}. The closest formulation, V-RFM, also combines a latent-conditioned velocity field with KL regularization~\cite{guo2025vrfm}, but studies noise-to-data generation. In O2MFlow, the latent instead indexes the appearance of an image-to-image transport: it is inferred from the target image, reused across sources, and held fixed to map real target images back to the source domain. Together, source-structure preservation, cross-source reuse, and backward translation define the image-to-image problem O2MFlow solves.

%% file: sec/3_preliminary.tex
\section{Preliminaries: Rectified Flow}\label{sec:standardRF}

RF models transport between source and target images $(\mathbf{X}_0, \mathbf{X}_1)$ by integrating a deterministic ODE defined by a learnable velocity field $v$~\cite{liu2023rectifiedflow}:
\begin{equation}
  \frac{\mathrm{d}\mathbf{X}_t}{\mathrm{d}t} =v(\mathbf{X}_t,t).  \label{eq:ode}
\end{equation}
To train $v$, we construct a straight-line interpolation:
\begin{equation}
  \mathbf{X}_t=t\mathbf{X}_1+(1-t)\mathbf{X}_0.
  \label{eq:linear}
\end{equation}
As shown in Fig.~\ref{fig:RF-vs-ours}(a), the velocity field $v$ is trained to match the constant direction $(\mathbf{X}_1-\mathbf{X}_0)$ along this interpolation path for all training pairs $\{ (\mathbf{X}_0, \mathbf{X}_1)\}$~\cite{liu2023rectifiedflow}:
\begin{equation}
  \min_{v}
  \int_{0}^{1}
  \mathbb{E}_{\mathbf{X}_0,\ \mathbf{X}_1}\Bigl[
      \bigl\|
        (\mathbf{X}_1-\mathbf{X}_0)
        -v(\mathbf{X}_t,t)
      \bigr\|_2^2
  \Bigr]\,
  \mathrm{d}t.
  \label{eq:rf_loss}
\end{equation}
\par

RF inference integrates the learned field from $\mathbf{X}_0$ to produce a deterministic target:
\begin{equation}
\widehat{\mathbf{X}}_1 = \mathbf{X}_0 + \int_0^{1} v(\mathbf{X}_t, t)\,\mathrm{d}t.
\label{eq:continuous-step}
\end{equation}
At the population optimum, squared regression gives $v^*(\mathbf{X}_t,t)=\mathbb E[\mathbf{X}_1-\mathbf{X}_0\mid\mathbf{X}_t,t]$, which averages the directions toward all targets compatible with $\mathbf{X}_t$. The deterministic transport yields one output per source, so representing alternative targets requires an additional variable such as an appearance latent.
\par

Deterministic ODE transports are {\em invertible} when the field is continuous in time and locally Lipschitz in the state, with solutions existing throughout $[0,1]$ in both directions. Under these conditions, integrating the same dynamics backward in time inverts the forward map. This property underpins invertible CNFs and ODE-based generative models~\cite{chen2018neuralode,grathwohl2019ffjord}, as shown in Fig.~\ref{fig:img2img}(b).
RF inherits this bidirectionality, enabling backward translation via time reversal without an explicit cycle-consistency loss~\cite{liu2023rectifiedflow}.

%% file: sec/4_proposedmethod.tex
\section{Method: O2MFlow}
\label{sec:proposed_method}

O2MFlow turns RF into a latent-indexed family of invertible transports, so that one model delivers both diverse and bidirectional translation. We formulate the paired task, define the conditions for reusable one-to-many transport, and introduce a model that can represent appearance variation while retaining fixed-latent invertibility. We then present the joint training objective and the role of KL regularization.

\subsection{Problem Formulation}
\label{sec:problem_formulation}

We consider paired training data $\{(\mathbf{X}_0^{(i)},\mathbf{X}_1^{(i)})\}_{i=1}^N$, with one observed target image $\mathbf{X}_1^{(i)}$ for each source $\mathbf{X}_0^{(i)}$. From these pairs, we learn a family of source-to-target mappings $T_{\mathbf{z}}$ indexed by an appearance latent $\mathbf{z}\sim p(\mathbf{z})$. Forward translation samples $\mathbf{z}$ and outputs $T_{\mathbf{z}}(\mathbf{X}_0)$, and backward translation maps a target-domain image to the source domain by $T_{\mathbf{z}}^{-1}$ with $\mathbf{z}$ inferred from that image. The family must therefore produce diverse, structure-preserving targets across latents, and each member must be invertible.

\subsection{Conditions for Shared Latent Transport}
\label{sec:shared_latent_transport}

\emph{Shared Latent Transport} specifies when the family $\{T_{\mathbf{z}}\}$ supports reusable one-to-many translation. It combines three conditions, illustrated in Fig.~\ref{fig:synchronized_latent} and formalized in Supplement~\ref{sec:proofs}: \emph{validity}, where every $T_{\mathbf{z}}(\mathbf{X}_0)$ is a plausible target preserving the structure of $\mathbf{X}_0$; \emph{nondegeneracy}, where different latents yield distinct outputs for the same source; and \emph{cross-source consistency}, where a fixed latent induces the same, source-adapted appearance variation for every source. Validity and nondegeneracy suffice for one-to-many generation, and cross-source consistency makes the variations reusable: an appearance learned from one pair transfers to sources never observed with it.

\input{fig/synchronized_latent}

Latent-conditioned RF parameterizes this family while retaining fixed-latent invertibility. A single velocity field shared across sources applies each latent to different inputs. To reuse a target-encoded latent across inputs, O2MFlow encodes the target without taking the source as an additional encoder input (Sec.~\ref{sec:translation}). This permits reuse but does not ensure cross-source consistency (Supplement~\ref{sec:proofs}); Fig.~\ref{fig:synchronized_latent} shows related appearance tendencies for the same latent across the displayed sources.

\subsection{Diverse and Bidirectional Translation}
\label{sec:translation}

O2MFlow represents this family with a shared RF velocity field $v_\theta(\mathbf{X}_t,t,\mathbf{z})$, conditioned on an appearance latent $\mathbf{z}$ with prior $p(\mathbf{z})=\mathcal N(0,\mathbf{I})$. The prior supplies latents for forward generation, and a target-only encoder $q_\phi(\mathbf{z}\mid \mathbf{X}_1)=\mathcal N(\bm{\mu}_\phi(\mathbf{X}_1),\operatorname{diag}(\bm{\sigma}_\phi^2(\mathbf{X}_1)))$ infers latents for training and backward translation. Figure~\ref{fig:RF-vs-ours}(b) summarizes both directions; architectural details are given in Sec.~\ref{subsec:exp_setup}.

\paragraph{Forward translation.}
For forward translation of $\mathbf{X}_0$, we sample $\mathbf{z}\sim p(\mathbf{z})$ and hold it fixed throughout integration from $t=0$ to $t=1$.
\begin{equation}
   \frac{\mathrm{d}\mathbf{X}_t}{\mathrm{d}t}
    =v_\theta(\mathbf{X}_t,t,\mathbf{z})
    \label{eq:latent_flow}
\end{equation}
This fixed-latent solve realizes $\widehat{\mathbf{X}}_1=T_{\mathbf{z}}(\mathbf{X}_0)$; nondegenerate dependence on $\mathbf{z}$ allows multiple target appearances for the same source.

\paragraph{Backward translation.}
For an observed target-domain image $\mathbf{X}_1$, the encoder estimates its latent $\mathbf{z}=\bm{\mu}_\phi(\mathbf{X}_1)$, and integrating the same ODE backward in time from $t=1$ yields $\widehat{\mathbf{X}}_0=T_{\mathbf{z}}^{-1}(\mathbf{X}_1)$. The inferred latent selects the inverse transport; paired-source recovery depends on both latent inference and learned correspondence.

Because each $T_{\mathbf{z}}$ is an ODE flow (Sec.~\ref{sec:standardRF}), backward translation needs neither a separate backward generator nor a cycle-consistency loss, and reusing the latent of a forward solve maps a generated target back to its source. We next describe how the encoder and transport are jointly learned from paired observations.

\subsection{KL-Regularized Transport Learning}
\label{sec:kl_learning}

During paired training, the encoder supplies latents that can convey appearance information to the velocity field. For each observed pair $(\mathbf{X}_0,\mathbf{X}_1)$, we draw
$\mathbf{z}\sim q_\phi(\mathbf{z}\mid \mathbf{X}_1)$ using $\mathbf{z}=\bm{\mu}_\phi(\mathbf{X}_1)+\bm{\sigma}_\phi(\mathbf{X}_1)\odot\bm{\epsilon}$, $\bm{\epsilon}\sim\mathcal N(0,\mathbf{I})$, and evaluate
the velocity field at the linear interpolation $\mathbf{X}_t$.
Following Lee et al.~\cite{lee2024improvingrf}, we apply a pseudo-Huber loss to the predicted target endpoint, using U-shaped time sampling. We denote this loss by $\mathcal L_{\mathrm{RF}}$; its explicit formulation is given in Supplement~\ref{sec:implementation}.

\input{fig/three_losses}

We jointly train $q_\phi$ and $v_\theta$ with RF regression and a KL penalty weighted by $\lambda_{\mathrm{KL}}$ (Fig.~\ref{fig:three-losses}):
\begin{equation}
    \mathcal L
    =\mathcal L_{\mathrm{RF}}
    +\lambda_{\mathrm{KL}}\mathcal L_{\mathrm{KL}},
    \label{eq:total_loss}
\end{equation}
where
\begin{equation}
    \mathcal L_{\mathrm{KL}}
    =\mathbb E_{\mathbf{X}_1}
    D_{\mathrm{KL}}\big(q_\phi(\mathbf{z}\mid \mathbf{X}_1)\,\|\,p(\mathbf{z})\big).
    \label{eq:loss_kl}
\end{equation}

KL regularization directly controls the source--latent distribution gap between training and generation. Let $P_{\mathrm{data}}$ denote the paired-data distribution and $P_0$ its source-image marginal. Sampling a pair and then a latent from its encoder defines the training joint $Q(\mathbf{X}_0,\mathbf{X}_1,\mathbf{z})=P_{\mathrm{data}}(\mathbf{X}_0,\mathbf{X}_1)q_\phi(\mathbf{z}\mid \mathbf{X}_1)$.
Its source--latent marginal is $Q(\mathbf{X}_0,\mathbf{z})$; generation instead samples source and prior latent independently from $P_0(\mathbf{X}_0)p(\mathbf{z})$.
For finite $\mathcal L_{\mathrm{KL}}$, the KL chain rule gives
\begin{equation}
  D_{\mathrm{KL}}\big(Q(\mathbf{X}_0,\mathbf{z})\,\|\,P_0(\mathbf{X}_0)p(\mathbf{z})\big)
  \le \mathcal L_{\mathrm{KL}}.
  \label{eq:source_latent_alignment}
\end{equation}
The difference between the right- and left-hand sides is the conditional mutual information $I_Q(\mathbf{X}_1;\mathbf{z}\mid \mathbf{X}_0)$ under $Q$ (derivation in Supplement~\ref{sec:kl_derivation}). For the empirical paired-data distribution with a unique target at each distinct source, this term vanishes and $\mathcal L_{\mathrm{KL}}$ equals the gap. Minimizing it aligns training and generation on those sources; the identity does not establish generalization to unseen sources.

The KL weight balances distribution alignment against target-dependent information in $\mathbf{z}$. A stronger penalty can suppress useful appearance information and reduce diversity. Section~\ref{sec:experiments} evaluates the resulting diversity--fidelity trade-off, cross-source reuse, and backward accuracy.

%% file: fig/synchronized_latent.tex
\begin{figure}[t]
\centering
\includegraphics[width=0.9\linewidth]{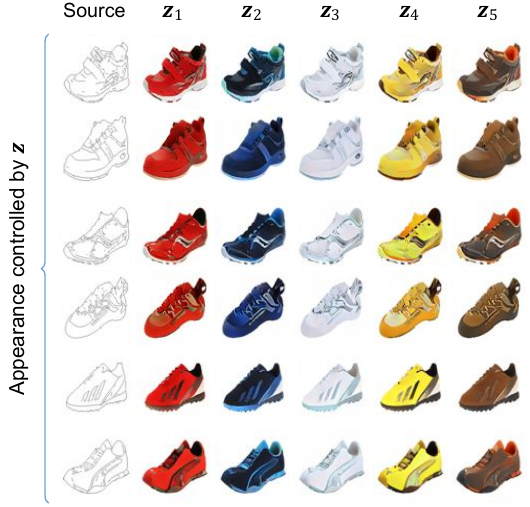}
\caption{Shared Latent Transport on edges2shoes. Reusing a target-encoded latent across sources produces related appearance tendencies while preserving source structure.}
\label{fig:synchronized_latent}
\end{figure}

%% file: fig/three_losses.tex
\begin{figure}[t]
  \centering
  \includegraphics[width=0.8\linewidth]{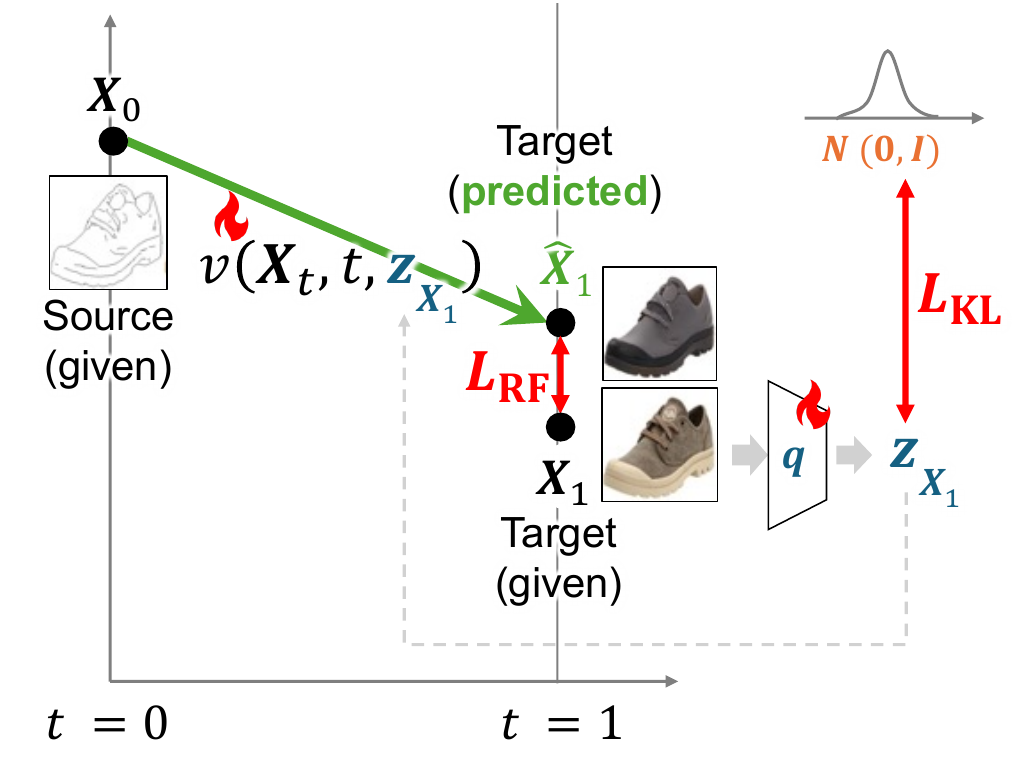}
  \caption{Training overview. RF regression fits paired target endpoints, and KL regularization aligns the target-encoded latent distribution with the prior. Flame symbols mark the jointly trained encoder and velocity field.}
  \label{fig:three-losses}
\end{figure}

%% file: sec/5_experiments.tex
\section{Experiments}\label{sec:experiments}

\subsection{Experimental Setup}\label{subsec:exp_setup}

We evaluate diverse forward generation and real-target backward recovery using the same model on edges2shoes, edges2handbags, and facades~\cite{isola2017pix2pix,zhu2017bicyclegan}. The edges2shoes and edges2handbags sketch-to-photo datasets contain 39,860/9,965/200 and 110,854/27,713/200 train/validation/test pairs, respectively. The architectural label-to-photo facades dataset contains 400/100/106 pairs in the same split order. In our experiments, all images were resized to $64\times64$. \par

A latent-conditioned U-Net~\cite{ronneberger2015unet} with self-attention is used to represent the velocity field $v_\theta(\mathbf{X}_t, t, \mathbf{z})$. The latent dimensionality is $d=16$; training uses posterior samples and forward evaluation uses independent standard-normal prior samples. To inject this condition into the U-Net, we use AdaIN-style feature modulation~\cite{huang2017adain,choi2020starganv2}. Specifically, a multilayer perceptron with four layers maps $\mathbf{z}$ to a 64-dimensional modulation vector.
The encoder $q_\phi$ is a convolutional neural network (CNN) trained from scratch. Both $q_\phi$ and $v_\theta$ are optimized with Eq.~\eqref{eq:total_loss}.
\par

Following Lee et al.~\cite{lee2024improvingrf}, we compute $\mathcal L_{\mathrm{RF}}$ with 64 time steps drawn from a U-shaped distribution. The endpoint pseudo-Huber objective is specified in Supplement~\ref{sec:implementation}. We train all models with the Adam optimizer (\(\beta_1=0.9, \beta_2=0.999\)) using a learning rate of \(1\times10^{-4}\) and a batch size of 64.
We use \(\lambda_{\mathrm{KL}}=0.02\) for all datasets. Training is stopped early based on the validation loss. Forward inference uses a fourth-order Runge--Kutta (RK4) solver with 50 steps over $[0,1]$~\cite{press2007numericalrecipes}.
\par

\subsection{Comparative Methods}\label{subsec:comparative method}

We compare deterministic, latent-variable, diffusion-bridge, and flow-based methods to assess structure preservation, diversity, and bidirectional translation. pix2pix~\cite{isola2017pix2pix} and CycleGAN~\cite{zhu2017cyclegan} provide single-output references for structure preservation and visual fidelity (Fig.~\ref{fig:img2img}). BicycleGAN~\cite{zhu2017bicyclegan} and DRIT++~\cite{lee2020dritpp} test one-to-many generation without an explicit inverse mapping (Fig.~\ref{fig:img2img}(c)). BBDM~\cite{li2023bbdm} and DDBM~\cite{zhou2024ddbm} add stochastic bridge-based forward translation. BDBM~\cite{kieu2025bdbm} provides a single-network bidirectional diffusion bridge, while SymmFlow~\cite{caetano2026symmflow} provides a flow-based bidirectional baseline.
% ----------------------------------------------------------------------------
\subsection{Evaluation Metrics}\label{subsec:metrics}

Forward evaluation of $K$ outputs per source measures distributional fidelity, output diversity, and paired-target agreement using Fréchet Inception Distance (FID)~\cite{heusel2017fid}, LPIPS-Div, and LPIPS-Dist, respectively. We average target-domain FID over the $K$ generations. Based on Learned Perceptual Image Patch Similarity (LPIPS)~\cite{zhang2018lpips}, LPIPS-Div is the average pairwise distance among the $K$ outputs for a source, and LPIPS-Dist is their average distance to the paired target $\mathbf{X}_1$. We use $K=5$ for one-to-many methods and $K=1$ for deterministic methods; LPIPS-Div is undefined for $K=1$.
\par

For backward translation, which produces a single reconstructed source image $\widehat{\mathbf{X}}_0$, we report two metrics: FID and LPIPS-Dist.
We compute FID in the source domain by comparing the set of reconstructed images $\{\widehat{\mathbf{X}}_0\}$ with the real source images $\{\mathbf{X}_0\}$. LPIPS-Dist is calculated as the LPIPS distance between the reconstructed image $\widehat{\mathbf{X}}_0$ and its paired ground-truth source image. \par

The three metrics are complementary. LPIPS-Dist measures agreement with the single observed reference and penalizes even valid alternative appearances, whereas FID measures marginal realism without source-specific content. We therefore read them jointly with LPIPS-Div as a diversity--fidelity trade-off.

\subsection{Same-Latent Round-Trip Validation}\label{subsec:roundtrip}
\input{fig/roundtrip}
Same-latent round trips assess the numerical reversibility of the learned transport. For each source $\mathbf{X}_0$, we sample five prior latents, compute $T_{\mathbf{z}}^{-1}(T_{\mathbf{z}}(\mathbf{X}_0))$ with the same $\mathbf{z}$ in both solves, and measure LPIPS to $\mathbf{X}_0$. Because the latent is known, this test isolates the transport from latent inference, which the real-target backward evaluation covers.
The round-trip LPIPS is 0.0010, 0.0053, and 0.0102 on
edges2shoes, edges2handbags, and facades, respectively, showing that the recovered source remains
very close to the original source after diverse forward generation.

\input{fig/qualiative_fig}

\subsection{Qualitative Evaluation}\label{subsec:qualitative}
O2MFlow varies target appearance while retaining source geometry in the forward examples (Figs.~\ref{fig:qualitative_fig}--\ref{fig:qualitative_facades}). In the method comparison (Fig.~\ref{fig:qualitative_fig}), the outputs retain shoe silhouettes and stripe patterns, bag handles and straps, and facade window rows while changing color and texture. In the displayed facades, pix2pix and CycleGAN show noisier window textures, whereas diffusion-bridge outputs retain more regular architectural patterns. Sampling different latents for a fixed source varies shoe and bag color and brightness while retaining the main outlines and component layout (Fig.~\ref{fig:qualitative_diverse}). The facades comparison further illustrates appearance variation within a shared label layout (Fig.~\ref{fig:qualitative_facades}): O2MFlow varies facade color and shading while keeping window placement broadly aligned with the input. BicycleGAN and DRIT++ show less regular window structures in the displayed samples, while DDBM shows relatively little appearance variation.

\input{fig/qualitative_diverse}
\input{fig/qualitative_facades}

O2MFlow recovers source-domain structure by reversing the same transport (Fig.~\ref{fig:qualitative_reverse}). The outputs reconstruct edge patterns for shoes and bags and colored region layouts for facades. In the displayed shoes, pix2pix produces faint or incomplete stripe patterns, and CycleGAN adds spurious edge responses; O2MFlow retains these patterns more clearly. CycleGAN and SymmFlow show noisy or diffuse structures in the displayed facade outputs, while BDBM and O2MFlow retain clearer label regions.

\input{fig/qualitative_reverse}

\subsection{Quantitative Evaluation}\label{subsec:quant_one_to_many}
\input{fig/tradeoff_pareto}
Forward results show a dataset-dependent diversity--fidelity trade-off (Table~\ref{tab:main_atob}). On edges2shoes and edges2handbags, O2MFlow produces higher LPIPS-Div than the diffusion- and flow-based baselines, with higher LPIPS-Dist and FID. On facades, O2MFlow is less diverse than BBDM and SymmFlow but more diverse than DDBM and BDBM, with lower LPIPS-Dist and FID than BBDM and BDBM.

O2MFlow achieves the lowest backward LPIPS-Dist on all three datasets and the lowest FID on edges2shoes and edges2handbags (Table~\ref{tab:main_btoa}). On facades, BDBM achieves lower FID, whereas O2MFlow has lower paired-source distance. These results support real-target source recovery with the forward transport model; same-latent numerical reversibility is assessed separately in Sec.~\ref{subsec:roundtrip}.
\input{table/main_atob}
\input{table/main_btoa}

\subsection{Analysis and Ablation}\label{subsec:analysis}
\paragraph{Synchronized Latent Grid.}
The provided synchronized grid illustrates cross-source appearance reuse (Fig.~\ref{fig:synchronized_latent}). Each column applies one latent, encoded from a reference target, to several source edge maps. The displayed outputs share related color tendencies within a column while retaining their respective source shapes. Together with appearance changes within a row, these observations connect the grid to the conditions in Sec.~\ref{sec:shared_latent_transport}. The evidence concerns the displayed examples and does not establish consistency for all unseen inputs.

\paragraph{Latent Interpolation.}
\input{fig/interpolation}
Latent interpolation examines appearance variation between two latent codes. For fixed source images, we interpolate two latents as $\mathbf{z}_\alpha=(1-\alpha)\mathbf{z}_a+\alpha\mathbf{z}_b$ with $\alpha\in[0,1]$. Appearance changes gradually while the source content is preserved (Fig.~\ref{fig:interpolation}), complementing the cross-source grid.

\paragraph{Effect of KL Regularization.}
\input{table/ablation_compact}
KL regularization improves forward fidelity over unregularized latent conditioning on all three datasets (Table~\ref{tab:ablation_compact}). Latent conditioning without KL generates diverse outputs but has higher forward FID and LPIPS-Dist than deterministic RF. Adding KL lowers both metrics while reducing LPIPS-Div. On edges2shoes, FID decreases from 83.28 to 50.86, LPIPS-Dist from 0.3187 to 0.1924, and LPIPS-Div from 0.3269 to 0.1776. KL therefore balances output diversity against fidelity, consistent with its distribution-alignment role in Eq.~\eqref{eq:source_latent_alignment}.

The deterministic baseline clarifies the distinction between marginal realism and paired-target agreement. The full model has lower forward FID but higher LPIPS-Dist than deterministic RF on all three datasets, while introducing measurable diversity. Thus, distance to the sole observed target alone does not determine the quality of one-to-many translation.

% Keep the first two lines of this paragraph together across columns.
{\clubpenalty=10000
Backward effects depend on the metric: KL lowers FID relative to latent conditioning alone on all three datasets, but LPIPS-Dist improves only on facades and increases slightly on the other two. Thus, the fidelity gains are more consistent in forward translation than in paired backward distance.
\par}
\input{fig/rf_only_qual}

\paragraph{Step Sensitivity.}
Forward and backward translation respond differently to the integration budget (Supplement~\ref{sec:historical_solver}). Reducing RK4 from 50 to 5 steps decreases NFE from 200 to 20, while forward FID on edges2shoes changes from 50.86 to 50.74. Backward FID increases from 22.57 to 26.93. Latency drops from 848.9 to 84.9\,ms per image (Supplement~\ref{sec:implementation}), enabling faster forward generation with little change in forward FID.

%% file: fig/roundtrip.tex
\begin{figure}[t]
  \centering
  \includegraphics[width=0.9\linewidth]{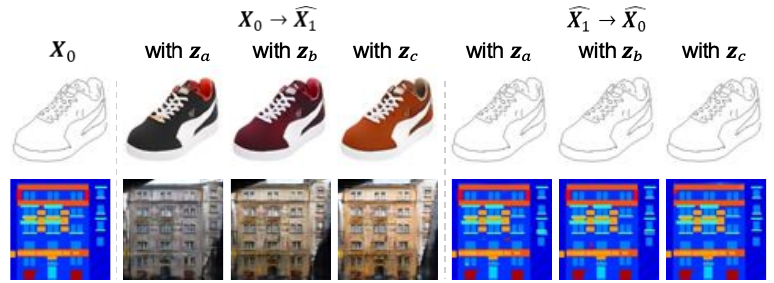}
   \caption{Same-latent round trip; the same fixed $\mathbf{z}$ is used in both directions. }
   \label{fig:roundtrip}
\end{figure}

%% file: fig/qualiative_fig.tex
\begin{figure}[t]
  \centering
  \includegraphics[width=1.0\linewidth]{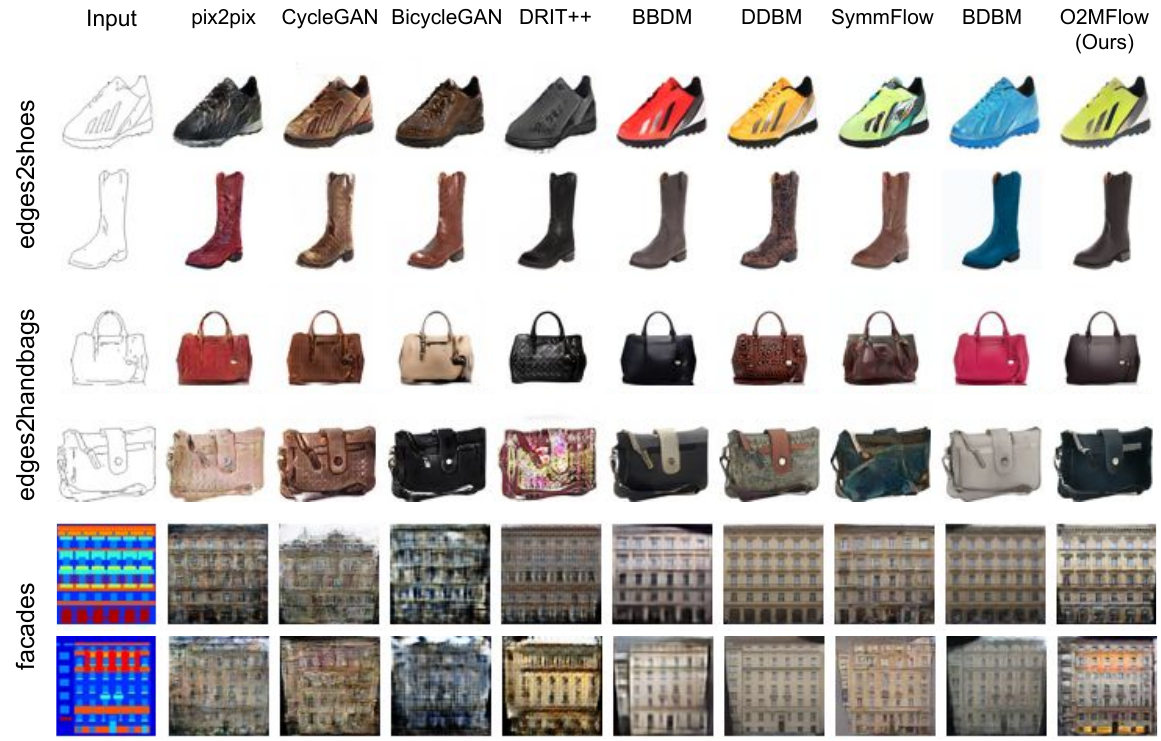}
  \caption{Qualitative comparison of forward translation on paired translation datasets.}
  \label{fig:qualitative_fig}
\end{figure}

%% file: fig/qualitative_diverse.tex
\begin{figure}[t]
  \centering
  \includegraphics[width=0.9\linewidth]{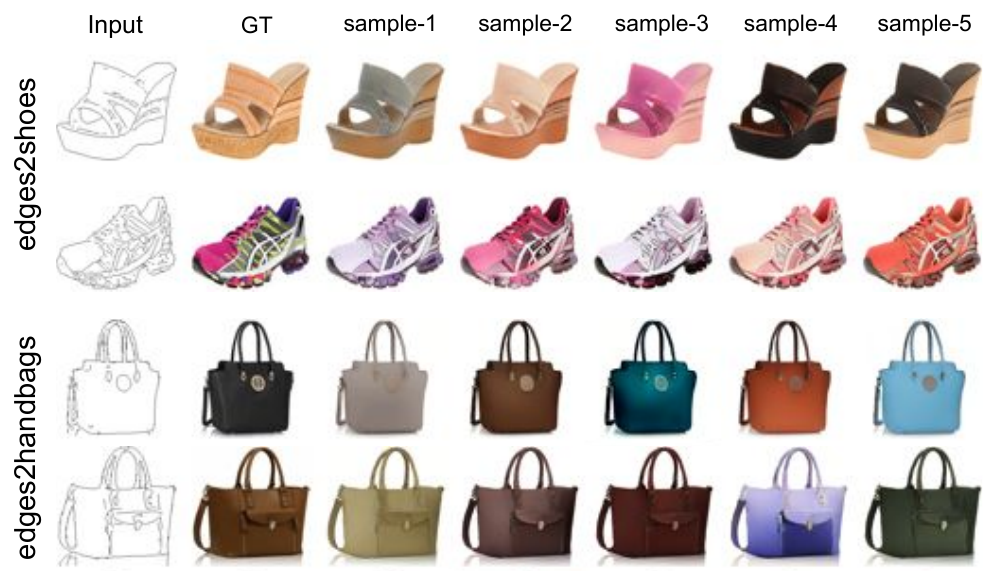}
  \caption{Forward translation results obtained by sampling different appearance latents.}
  \label{fig:qualitative_diverse}
\end{figure}

%% file: fig/qualitative_facades.tex
\begin{figure}[t]
  \centering
  \includegraphics[width=1.0\linewidth]{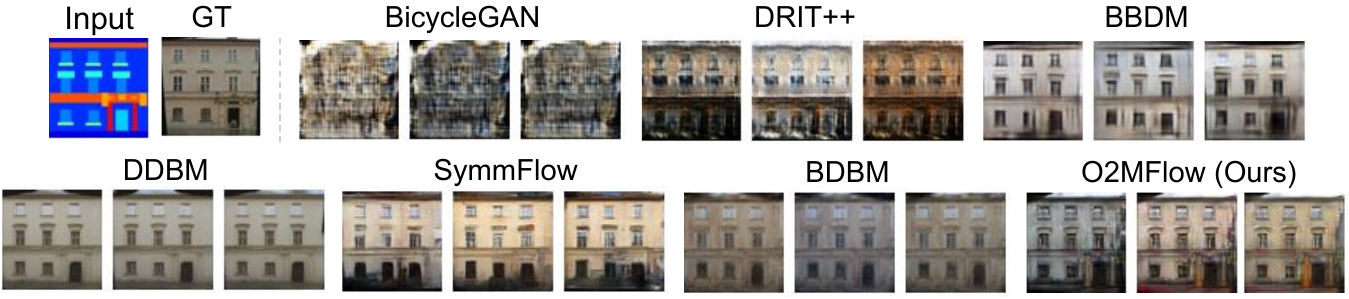}
  \caption{Comparison of output diversity on the facades dataset.}
  \label{fig:qualitative_facades}
\end{figure}

%% file: fig/qualitative_reverse.tex
\begin{figure}[t]
  \centering
  \includegraphics[width=1.0\linewidth]{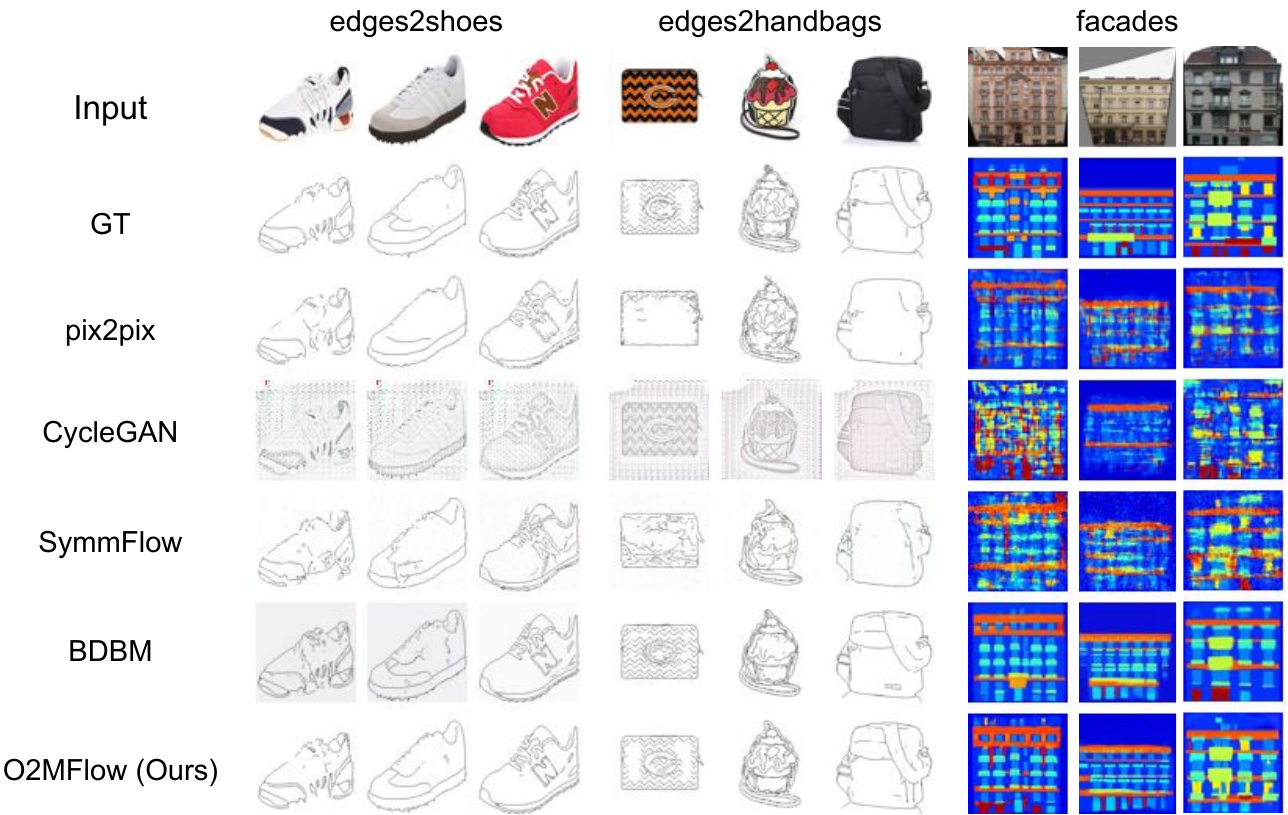}
  \caption{Backward translation results on paired translation datasets. We compare our single-model backward translation against pix2pix trained separately for the backward direction, CycleGAN, SymmFlow, and BDBM.}
  \label{fig:qualitative_reverse}
\end{figure}

%% file: fig/tradeoff_pareto.tex
\begin{figure*}[t]
  \centering
  \includegraphics[width=0.9\textwidth]{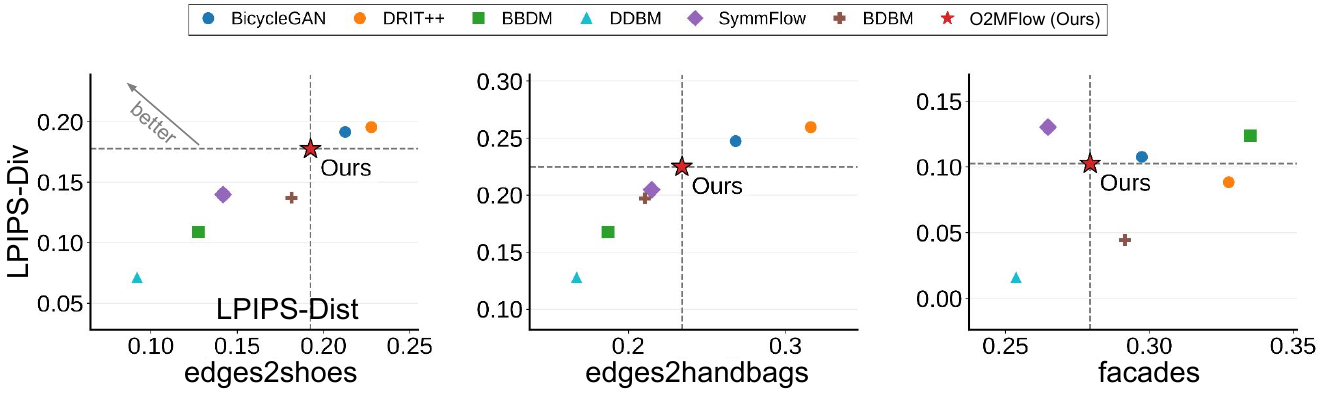}
  \caption{Forward diversity--fidelity trade-off on three datasets. The upper-left region combines higher LPIPS-Div with lower LPIPS-Dist.
 }
  \label{fig:tradeoff_pareto}
\end{figure*}

%% file: table/main_atob.tex
\begin{table}[t]
  \centering
  \caption{Quantitative comparison on forward translation.}
  \label{tab:main_atob}
  \footnotesize
  \renewcommand{\arraystretch}{1.0}
  \setlength{\tabcolsep}{4.5pt}
  \begin{tabular}{lccc}
    \toprule
    \textbf{Dataset}/Method & LPIPS-Div
    $\uparrow$ & LPIPS-Dist $\downarrow$ & FID $\downarrow$ \\
    \midrule
    \multicolumn{4}{l}{\textbf{edges2shoes}} \\
    pix2pix            & --     & 0.1630 &  99.71 \\
    CycleGAN           & --     & 0.2050 & 108.88 \\
    BicycleGAN         & 0.1914 & 0.2130 &  58.51 \\
    DRIT++             & 0.1954 & 0.2280 &  50.18 \\
    BBDM               & 0.1089 & 0.1280 &  37.44 \\
    DDBM               & 0.0713 & 0.0920 &  32.44 \\
    SymmFlow           & 0.1397 & 0.1420 &  38.17 \\
    BDBM               & 0.1371 & 0.1815 &  44.08 \\
    \rowcolor{cvprblue!10}
    O2MFlow            & 0.1776 & 0.1924 &  50.86 \\
    \midrule
    \multicolumn{4}{l}{\textbf{edges2handbags}} \\
    pix2pix            & --     & 0.2490 &  96.29 \\
    CycleGAN           & --     & 0.2789 &  95.43 \\
    BicycleGAN         & 0.2473 & 0.2687 &  73.02 \\
    DRIT++             & 0.2595 & 0.3164 &  86.70 \\
    BBDM               & 0.1675 & 0.1875 &  60.33 \\
    DDBM               & 0.1279 & 0.1671 &  58.78 \\
    SymmFlow           & 0.2046 & 0.2152 &  65.79 \\
    BDBM               & 0.1972 & 0.2105 &  60.16 \\
    \rowcolor{cvprblue!10}
    O2MFlow            & 0.2248 & 0.2340 &  69.13 \\
    \midrule
    \multicolumn{4}{l}{\textbf{facades}} \\
    pix2pix            & --     & 0.3092 & 252.78 \\
    CycleGAN           & --     & 0.3470 & 212.38 \\
    BicycleGAN         & 0.1076 & 0.2976 & 326.87 \\
    DRIT++             & 0.0883 & 0.3279 & 222.09 \\
    BBDM               & 0.1230 & 0.3356 & 147.28 \\
    DDBM               & 0.0161 & 0.2537 &  93.94 \\
    SymmFlow           & 0.1302 & 0.2651 & 107.70 \\
    BDBM               & 0.0442 & 0.2915 & 113.64 \\
    \rowcolor{cvprblue!10}
    O2MFlow            & 0.1024 & 0.2794 & 109.78 \\
    \bottomrule
    \end{tabular}
\end{table}

%% file: table/main_btoa.tex
\begin{table}[t]
\centering
\caption{Quantitative comparison on backward translation.}
\label{tab:main_btoa}
\footnotesize
\renewcommand{\arraystretch}{0.9}
\setlength{\tabcolsep}{5pt}
\begin{tabular}{lcc}
\toprule
\textbf{Dataset}/Method & LPIPS-Dist $\downarrow$ & FID $\downarrow$ \\
\midrule
\multicolumn{3}{l}{\textbf{edges2shoes}} \\
pix2pix (baseline) & 0.0501 &  36.68 \\
CycleGAN          & 0.2309 & 197.79 \\
SymmFlow          & 0.0834 & 109.98 \\
BDBM              & 0.0702 &  43.97 \\
\rowcolor{cvprblue!10}
O2MFlow           & 0.0287 &  22.57 \\
\midrule
\multicolumn{3}{l}{\textbf{edges2handbags}} \\
pix2pix (baseline) & 0.0907 &  46.00 \\
CycleGAN          & 0.1979 & 188.96 \\
SymmFlow          & 0.1165 &  82.27 \\
BDBM              & 0.0628 &  35.21 \\
\rowcolor{cvprblue!10}
O2MFlow           & 0.0532 &  32.29 \\
\midrule
\multicolumn{3}{l}{\textbf{facades}} \\
pix2pix (baseline) & 0.2022 & 188.70 \\
CycleGAN          & 0.2620 & 194.78 \\
SymmFlow          & 0.1971 & 261.15 \\
BDBM              & 0.2040 &  72.18 \\
\rowcolor{cvprblue!10}
O2MFlow           & 0.1517 &  74.68 \\
\bottomrule
\end{tabular}
\end{table}

%% file: fig/interpolation.tex
\begin{figure}[t]
  \centering
    \includegraphics[width=0.9\linewidth]{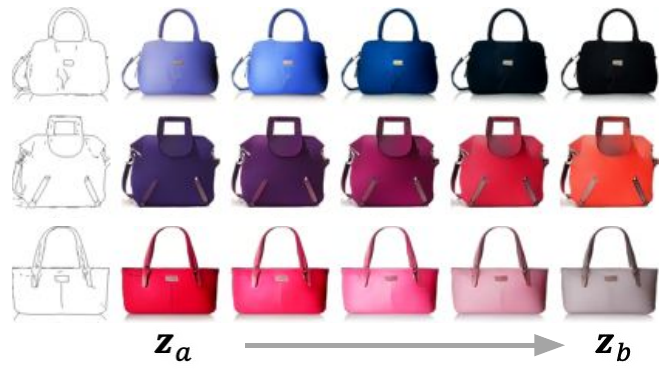}
   \caption{Latent interpolation: appearance changes smoothly while the source content is preserved.}
   \label{fig:interpolation}
\end{figure}

%% file: table/ablation_compact.tex
\begin{table}[t]
\centering
\caption{Ablation study of the proposed components on the edges2shoes, edges2handbags, and facades datasets. Latent denotes appearance-latent conditioning; KL denotes regularization of the target-encoded latent distribution toward the prior by $\mathcal L_{\mathrm{KL}}$ in Eq.~\eqref{eq:loss_kl}. Checkmarks indicate enabled components.}
\label{tab:ablation_compact}
\setlength{\tabcolsep}{3pt}
\renewcommand{\arraystretch}{1.15}
\resizebox{\columnwidth}{!}{%
\begin{tabular}{ccccccc}
\toprule
\multicolumn{2}{c}{Components} & \multicolumn{3}{c}{Forward} & \multicolumn{2}{c}{Backward} \\
\cmidrule(lr){1-2}\cmidrule(lr){3-5}\cmidrule(lr){6-7}
Latent & KL & LPIPS-Div $\uparrow$ & LPIPS-Dist $\downarrow$ & FID $\downarrow$ & LPIPS-Dist $\downarrow$ & FID $\downarrow$ \\
\midrule
\multicolumn{7}{l}{\textbf{edges2shoes}} \\
-- & -- & -- & 0.1013 & 54.02 & 0.0277 & 23.09 \\
\cmark & -- & 0.3269 & 0.3187 & 83.28 & 0.0281 & 23.36 \\
\rowcolor{cvprblue!10}
\cmark & \cmark & 0.1776 & 0.1924 & 50.86 & 0.0287 & 22.57 \\
\midrule
\multicolumn{7}{l}{\textbf{edges2handbags}} \\
-- & -- & -- & 0.1657 & 74.16 & 0.0546 & 32.81 \\
\cmark & -- & 0.4478 & 0.4287 & 130.04 & 0.0521 & 32.49 \\
\rowcolor{cvprblue!10}
\cmark & \cmark & 0.2248 & 0.2340 & 69.13 & 0.0532 & 32.29 \\
\midrule
\multicolumn{7}{l}{\textbf{facades}} \\
-- & -- & -- & 0.2572 & 119.33 & 0.1482 & 67.09 \\
\cmark & -- & 0.1996 & 0.3177 & 140.49 & 0.1566 & 74.97 \\
\rowcolor{cvprblue!10}
\cmark & \cmark & 0.1024 & 0.2794 & 109.78 & 0.1517 & 74.68 \\
\bottomrule
\end{tabular}}
\end{table}

%% file: fig/rf_only_qual.tex
\begin{figure}[t]
  \centering
  \includegraphics[width=\linewidth]{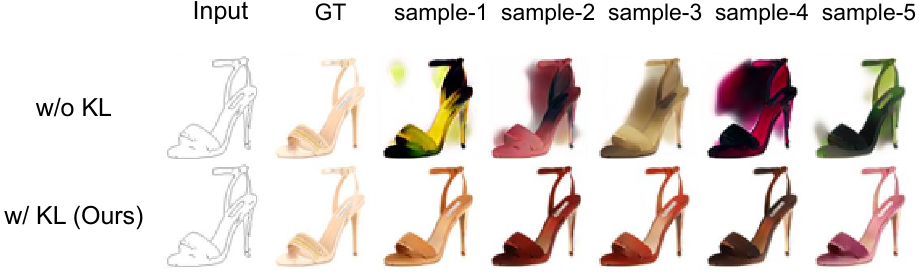}
  \caption{Effect of KL regularization on edges2shoes. The top and bottom rows show samples without and with KL, respectively; Table~\ref{tab:ablation_compact} reports the quantitative ablation.}
  \label{fig:rf_only_qual}
\end{figure}

%% file: sec/6_conclusion.tex
\section{Conclusion}

We proposed O2MFlow for learning one-to-many paired image translation from one observed target per source while supporting backward translation with the same transport model. An appearance latent conditions the Rectified Flow dynamics to represent appearance variation, while fixed-latent invertibility allows backward translation via time-reversed integration. We formalized cross-source variation reuse as Shared Latent Transport and showed that KL regularization directly controls the source--latent distribution gap between training and generation. Among evaluated methods, O2MFlow achieves the lowest backward LPIPS-Dist on all three benchmarks and the lowest backward FID on two, alongside dataset-dependent forward diversity--fidelity trade-offs. The ablation shows that KL improves forward fidelity over unregularized latent conditioning while retaining diversity.

Quantitative evaluation is limited to $64\times64$ images, with $128\times128$ examples providing qualitative evidence only. Future work includes scaling fixed-latent invertible transport to higher resolutions with scalable RF architectures and reducing ODE evaluations through solver optimization or distillation while maintaining backward accuracy. Structured, interpretable appearance latents could further enhance controllability through reference guidance or attribute control.

%% file: sec/8_llm_usage.tex
\section*{LLM Usage Statement}
OpenAI Codex and Claude Code both assisted with manuscript editing, translation,
review, related-work searches, bibliographic metadata verification, and code
generation.

%% file: sec/x_supp.tex
\clearpage
\maketitlesupplementary
\appendix

This document supplements ``One-to-Many: Diverse and Bidirectional Image-to-Image Translation via Rectified Flow.'' It formalizes Shared Latent Transport and the KL analysis (Sec.~\ref{sec:proofs}), details the implementation (Sec.~\ref{sec:implementation}), and reports additional evaluations (Sec.~\ref{sec:additional_eval}). References are shared with the main paper.

\section{Shared Latent Transport and KL Analysis}\label{sec:proofs}
\subsection{Interpretation of Shared Latent Transport}
This section formalizes the three conditions of Sec.~\ref{sec:shared_latent_transport} and shows that they let variations be recombined across sources. Let $P_0$ be the source-image distribution and $p$ the latent prior. The task-valid set $\mathcal V(\mathbf{x})$ contains the targets that preserve the required content of a source $\mathbf{x}$, and the descriptor $a(\mathbf{x},\mathbf{y})$ measures the task-relevant variation of an output $\mathbf{y}$ independently of the learned encoder; both are evaluation concepts, not additional networks or losses. Assuming measurable transports and descriptors, validity and cross-source consistency require
\begin{equation}
\begin{gathered}
T_{\mathbf{z}}(\mathbf{x})\in\mathcal V(\mathbf{x}),\\
a(\mathbf{x},T_{\mathbf{z}}(\mathbf{x}))
=a(\mathbf{x}',T_{\mathbf{z}}(\mathbf{x}'))
\end{gathered}
\label{eq:shared_latent_property}
\end{equation}
almost everywhere under $P_0\otimes P_0\otimes p$, and nondegeneracy requires, for $P_0$-almost every $\mathbf{x}$,
\begin{equation}
\Pr_{\mathbf{z},\mathbf{z}'\stackrel{\mathrm{iid}}\sim p}
\left[a(\mathbf{x},T_{\mathbf{z}}(\mathbf{x}))\ne a(\mathbf{x},T_{\mathbf{z}'}(\mathbf{x}))\right]>0.
\end{equation}
Validity places outputs in $\mathcal V(\mathbf{x})$ almost surely, and nondegeneracy yields distinct descriptors, and hence distinct outputs, with positive probability. Consistency equates the descriptors for almost every source pair and shared latent, allowing variations to recombine across sources under the stated assumptions. This recombination follows directly from the conditions, which leaves the question of how a model learns them from finite pairs.

Parameter sharing and prior alignment alone do not determine what a latent means for each source. For two source types $\mathbf{x}_A,\mathbf{x}_B$ and a symmetric Gaussian latent, the descriptor maps $a(\mathbf{x}_A,T_{\mathbf{z}}(\mathbf{x}_A))=\mathbf{z}$ and $a(\mathbf{x}_B,T_{\mathbf{z}}(\mathbf{x}_B))=-\mathbf{z}$ share the same attribute marginal, yet the same latent has opposite meanings. O2MFlow uses a target-only encoder (Sec.~\ref{sec:translation}) to obtain latents that can be applied to different sources. This permits reuse but does not resolve the semantic ambiguity above; Fig.~\ref{fig:synchronized_latent} tests consistency on concrete sources and latents.

\subsection{Source--latent distribution gap under KL regularization}
\label{sec:kl_derivation}
The KL analysis compares target-encoded training latents with independently sampled generation latents. With $P_{\mathrm{data}}$ denoting the paired-data distribution, define the training joint
\begin{equation}
Q(\mathbf{x}_0,\mathbf{x}_1,\mathbf{z})=P_{\mathrm{data}}(\mathbf{x}_0,\mathbf{x}_1)q_\phi(\mathbf{z}\mid \mathbf{x}_1)
\end{equation}
and the generation distribution $R(\mathbf{x}_0,\mathbf{z})=P_0(\mathbf{x}_0)p(\mathbf{z})$, where $P_0$ is the source marginal of $P_{\mathrm{data}}$ and $p$ is the latent prior. Write $Q_{\mathbf{X}_0,\mathbf{z}}$ for the source--latent marginal of $Q$. Assume absolute continuity with respect to the corresponding reference distributions and finite $\mathcal L_{\mathrm{KL}}$. The chain rule gives
\begin{equation}
\begin{aligned}
\mathcal L_{\mathrm{KL}}
&=D_{\mathrm{KL}}(Q\,\|\,P_{\mathrm{data}}p)\\
&=D_{\mathrm{KL}}(Q_{\mathbf{X}_0,\mathbf{z}}\,\|\,P_0p)\\
&\quad+I_Q(\mathbf{X}_1;\mathbf{z}\mid \mathbf{X}_0).
\end{aligned}
\label{eq:kl_decomposition}
\end{equation}
The $\mathbf{X}_0,\mathbf{X}_1$ marginal of $Q$ is $P_{\mathrm{data}}$, so the residual is the conditional mutual information shown above. Its nonnegativity proves the main-paper inequality; this standard KL chain-rule identity, interpreted for the paired image-to-image sampling mismatch, holds for any RF regression loss. Under an empirical distribution with a unique target at each distinct source, $I_Q(\mathbf{X}_1;\mathbf{z}\mid \mathbf{X}_0)=0$ and the inequality is an equality, so $\mathcal L_{\mathrm{KL}}$ measures exactly the gap on the training sources.

Pinsker's inequality bounds the total variation (TV) between the training and generation source--latent distributions:
\begin{equation}
\operatorname{TV}(Q_{\mathbf{X}_0,\mathbf{z}},P_0p)\le\sqrt{\mathcal L_{\mathrm{KL}}/2}.
\end{equation}
For any measurable $f(\mathbf{x},\mathbf{z})\in[0,1]$, such as the indicator of a latent region for a set of sources, expectations under training and generation differ by at most this quantity. The bound concerns source--latent pairs rather than image-quality metrics such as FID or LPIPS, which Sec.~\ref{subsec:analysis} evaluates directly.

\subsection{Fixed-latent inversion}
Fixed-latent inversion is exact in continuous time. If $v_\theta(\cdot,t,\mathbf{z})$ is locally Lipschitz in the image state, continuous in time, and solutions exist on the whole interval in both directions, uniqueness defines an inverse flow on its image, so $T_{\mathbf{z}}^{-1}(T_{\mathbf{z}}(\mathbf{x}))=\mathbf{x}$; finite-step numerical solvers approximate this identity. For a real target $\mathbf{y}$, the recovery $T_{\bm{\mu}_\phi(\mathbf{y})}^{-1}(\mathbf{y})$ additionally depends on latent inference and the learned paired correspondence. The experiments therefore evaluate same-latent round trips (Sec.~\ref{subsec:roundtrip}) separately from real-target backward translation (Sec.~\ref{subsec:quant_one_to_many}).

\section{Implementation Details}\label{sec:implementation}
\subsection{Encoder and joint optimization}
The encoder is a four-layer strided CNN with $4\times4$ convolutions and LeakyReLU activations, followed by a Gaussian latent head that outputs the parameters of $q_\phi(\mathbf{z}\mid \mathbf{X}_1)=\mathcal N(\bm{\mu}_\phi(\mathbf{X}_1),\operatorname{diag}(\bm{\sigma}_\phi^2(\mathbf{X}_1)))$. The encoder and velocity field are optimized jointly with $\mathcal L_{\mathrm{RF}}$ (Eq.~\eqref{eq:loss_rf_endpoint}) and $\mathcal L_{\mathrm{KL}}$; O2MFlow uses no latent-regression branch or separate backward generator.

\subsection{Implemented endpoint regression}
Following Lee et al.~\cite{lee2024improvingrf}, $\mathcal L_{\mathrm{RF}}$ applies a pseudo-Huber penalty to the predicted target endpoint. For a training interpolation $\mathbf{X}_t$ and posterior sample $\mathbf{z}$, the endpoint prediction is
\begin{equation}
\widehat{\mathbf{X}}_1^{(t)}=\mathbf{X}_t+(1-t)v_\theta(\mathbf{X}_t,t,\mathbf{z}),
\label{eq:x1_hat}
\end{equation}
and the objective is
\begin{equation}
\mathcal L_{\mathrm{RF}}
=\mathbb E_{Q,\pi}\big[\rho_\delta(\widehat{\mathbf{X}}_1^{(t)}-\mathbf{X}_1)\big],
\label{eq:loss_rf_endpoint}
\end{equation}
where the expectation is over the training joint $Q$ and the U-shaped time distribution $\pi$, and $\rho_\delta$ is the pseudo-Huber penalty. We set $\lambda_{\mathrm{KL}}=0.02$.

Endpoint regression differs from the squared velocity regression of Eq.~\eqref{eq:rf_loss}: even a squared endpoint error carries a $(1-t)^2$ factor relative to the velocity error, and pseudo-Huber further reshapes the penalty. The KL analysis in Sec.~\ref{sec:kl_derivation} holds for either objective, because it involves only the encoder and the prior.

\subsection{Training/inference cost and NFE}
We report training and inference costs on a single RTX A6000. Training on edges2shoes takes about 12.5h. At inference, 50 RK4 steps use 200 function evaluations (NFE=200), matching the NFE budget of all diffusion- and flow-based baselines. For a runtime comparison with BBDM, the measured latency is 848.9\,ms per image for our model and 1374.9\,ms for BBDM; reducing our solver to 5 steps gives 84.9\,ms with nearly unchanged forward FID, at the expense of backward accuracy (Table~\ref{tab:step_sensitivity}).

\section{Additional Evaluation}\label{sec:additional_eval}
\subsection{Diversity beyond color}
Gray-LPIPS-Div measures output variation after grayscale conversion. Comparing it with LPIPS-Div assesses how much variation remains when chromatic differences are removed.
The Gray-LPIPS-Div is 0.092, 0.117, and 0.081 on edges2shoes, edges2handbags, and facades, respectively, retaining 52\%, 52\%, and 79\% of the original LPIPS-Div. This suggests that the diversity is not solely due to chromatic changes, but also includes non-color appearance cues such as texture, shading, and material-like variation. Since paired I2I should preserve the input-defined geometry, our goal is not arbitrary geometric deformation but meaningful appearance diversity under fixed structure.

\subsection{Scalability beyond \texorpdfstring{$64\times64$}{64 by 64}}
\begin{figure}[h]
\centering
\includegraphics[width=\linewidth]{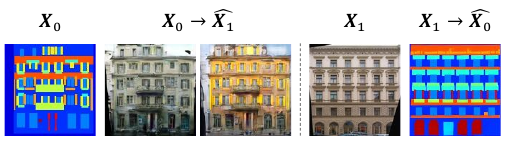}
\caption{$128\times128$ results of O2MFlow on facades.}
\label{fig:facades128}
\end{figure}

The $128\times128$ facades examples in Fig.~\ref{fig:facades128} extend the $64\times64$ main-benchmark setting. Higher resolution increases the image-state dimensionality and solver cost, which makes a learned latent image space a natural extension.

\subsection{Solver step sensitivity}\label{sec:historical_solver}
\input{table/step_sensitivity}
Increasing RK4 steps mainly improves backward FID in this comparison. Table~\ref{tab:step_sensitivity} reports results on edges2shoes, where Fwd. and Bwd. denote forward and backward translation. Increasing the steps from 5 to 50 reduces Bwd. FID from 26.93 to 22.57, whereas forward translation changes only slightly: Fwd. FID stays nearly constant (50.69--50.86), and Fwd. LPIPS-Div increases from 0.1710 to 0.1776. These measurements suggest that fewer steps may suffice for forward translation on this dataset, while backward accuracy benefits from additional integration steps.

%% file: table/step_sensitivity.tex
\begin{table}[t]
\centering
\caption{RK4 step sensitivity on edges2shoes.}
\label{tab:step_sensitivity}
\small
\setlength{\tabcolsep}{5pt}
\begin{tabular}{@{}cccc@{}}
\toprule
Steps & Fwd. LPIPS-Div~$\uparrow$ & Fwd. FID~$\downarrow$ & Bwd. FID~$\downarrow$ \\
\midrule
5  & 0.1710 & 50.74 & 26.93 \\
10 & 0.1721 & 50.74 & 23.21 \\
25 & 0.1748 & \textbf{50.69} & 22.66 \\
50 & \textbf{0.1776} & 50.86 & \textbf{22.57} \\
\bottomrule
\end{tabular}
\end{table}